\documentclass[sigconf]{acmart}
\usepackage[utf8]{inputenc}

\setcopyright{acmcopyright}
\copyrightyear{2022}
\acmYear{2022}
\acmDOI{XXXXXXX.XXXXXXX}

\acmConference[TRAITS-HRI 2022]{The Road to a successful HRI: AI, Trust and ethicS}{March 11,
  2022}{Sapporo, Japan}

\title{Introducing RISK}
\subtitle{Enabling Hypothetical Reasoning For Transparency in Robotics}

\author{Christopher D. Wallbridge}
\email{wallbridgec@cardiff.ac.uk}
\orcid{0000-0001-9468-122X}
\affiliation{%
  \institution{IROHMS}
  \streetaddress{Cardiff University}
  \city{Cardiff}
  \country{Wales}
  \postcode{CF10 3AT}
}

\author{Qiyuan Zhang}
\email{zhangq47@cardiff.ac.uk}
\affiliation{%
  \institution{IROHMS}
  \streetaddress{Cardiff University}
  \city{Cardiff}
  \country{Wales}
  \postcode{CF10 3AT}
}

\begin{document}

\maketitle

\section{Introduction}

One barrier to more prevalent use of robots and AI is the trust we put in these systems. Especially cases where a robot is acting in safety critical situations it is necessary that we feel able to trust and understand the robot. Advances in machine and deep learning technologies, while greatly increasing the capability of robots, have come at the cost of scrutability. The models that are created by these techniques are often considered a `black box', with an input coming in one side and an output coming out at the end, without a real understanding of how that process has been achieved.

These deep learning techniques are based on neural networks, which we believe is the way the human mind works. If we look at the example of human perception, we see that machines that have used deep learning techniques can be fooled by visual illusions~\cite{watanabe2018illusory}, which would seem to suggest that the network is processing information in a similar way to a person. In many ways the human mind is as much of a black box as any deep learning techniques. However we are able to trust other people with safety critical processes. We should be able to develop processes that allow us to trust a robot for the same reason we would trust person~\cite{bryson2019society}.

If we are investigating an incident, involving a person being responsible, a wide range of factors will be considered. From this we build a list of items that resulted in the accident based on evidence gathered from the person responsible, witnesses, footage of the incident, and other forms of data. Having established causes of the incident we can make changes to prevent future accidents:
\begin{itemize}
    \item Re-training individuals involved in the incident.
    \item Improve perception.
    \item Improve procedures.
\end{itemize}
These corrective actions are just as valid for a robotic system that has been involved in an accident. A deep network can be re-trained with more emphasis on conditions that led to the accident, even using data from the accident. Where a person might be given glasses to fix a fault in their vision a robot can be upgraded with better sensors. Finally if the procedures themselves are at fault then the robot's algorithms can be changed. However robots currently lack the tools to be able to reason about and explain their decisions.

Introduced here is a system for Rapid Internal Simulation of Knowledge (RISK) so that a robot may simulate its own actions. The principles behind this framework:
\begin{itemize}
    \item Real-Time Operation - The actions need to be simulated in a way that allows the robot to act in a timely manner.
    \item On Board - These simulations should not require a large amount of external processing power, but should be part of the robot's regular operation, allowing it to simulate its own actions. This differs from many simulation frameworks that allow programmers to simulate actions they have programmed before running on the robot.
    \item Human Understandable - The simulations that are created should be replayable and understandable without expert knowledge. This allows judgements to be made on the robot's actions.
\end{itemize}

\section{Design}

At it's core RISK uses Underworlds~\cite{lemaignan2018underworlds} to provide a light weight frame work to allow spatial and temporal models of the environment, and can create multiple versions of a scenario called worlds. For instance one world might be perspective according to a particular observer. These worlds can be displayed in real time, which can increase transparency to observers~\cite{wortham2017improving}. RISK builds upon this capability to look at hypothetical situations.

Results from a machine learned system can be tested by projecting into a future world. These simulations are recorded, and can be played back after an incident to improve transparency to observers. By comparing the results of the hypothetical world to what actually happened should make clear, if there was any faults in the training, faults in perception of entity or objects, or even if the simulations themselves are flawed. This kind of reasoning process, comparing what could have happened to what did happen, is a form of counterfactual thinking~\cite{kahneman1986norm, roese1997counterfactual}. Counterfactual thinking is a natural form of reasoning after incidents have a negative outcome, and by generating natural language explanations based on this kind of reasoning may also serve to increase the transparency of a robot's actions.

\section{Example Scenario}

\begin{figure}[h]
    \centering
    \includegraphics[width=\columnwidth]{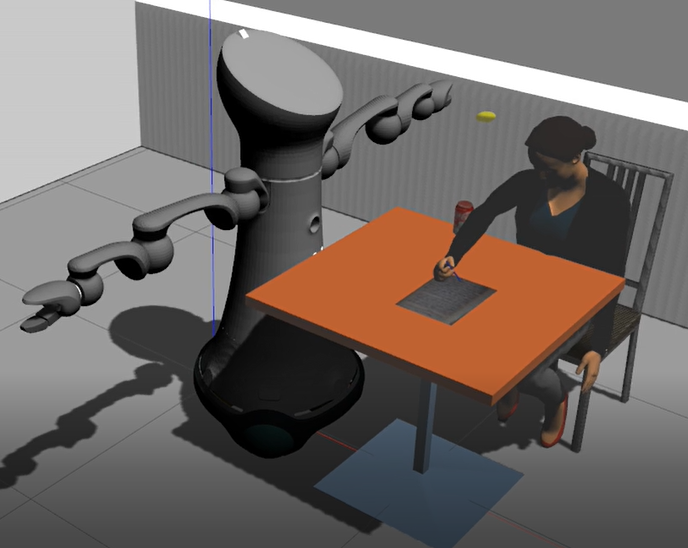}
    \caption[Image showing the example scenario, with a person writing at a table, there is a coke can also on the table. The robot is in a position to be able to remove the cola can.]{Image showing the example scenario, with a person writing at a table, there is a coke can also on the table. The robot is in a position to be able to remove the coke can. The example scenario was programmed in Gazebo.}
    \label{fig:ExScenario}
\end{figure}

\begin{figure}[h]
    \centering
    \includegraphics[width=\columnwidth]{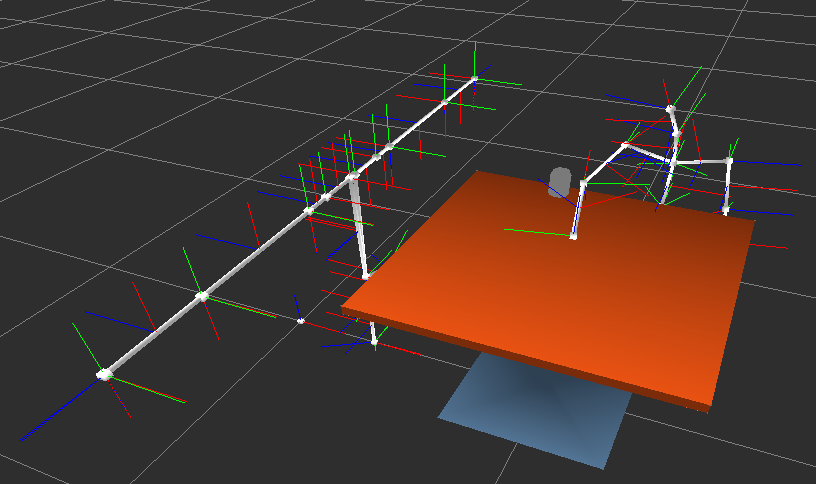}
    \caption[A semantic representation of the scenario, showing the current position of the person, the robot and the can.]{A semantic representation of the scenario, showing the current position of the person, the robot and the can.}
    \label{fig:SemRep}
\end{figure}

An example scenario was used to develop the RISK framework. The scenario had a person writing at a table. A can was placed at the person's arm, in a position where it could be knocked over. The robot was positioned to intervene (see figure~\ref{fig:ExScenario}). Using a semantic representation of the person (see figure~\ref{fig:SemRep}) the robot is able to calculate whether the person can accidentally knock over the can --the hypothetical situation-- and then act based on this information.

By manipulating parameters such as position of the coke and the representation of the person's model, it is possible to make an accident appear possible or impossible to the robot. A study needs to be conducted to ensure that this makes the reasons for the robots decision scrutable to a person without expert knowledge.

\section{Ongoing Work}

There are a number of still ongoing developments for the RISK system:
\begin{itemize}
    \item \emph{User study} - One of the main objectives is to ensure that the robot can be understood. The study will be conducted using recordings of the simulations conducted by the robot.
    \item \emph{Transfer to a real robot} - Currently much of the implementation has been programmed using Gazebo due to current working conditions. Simulating a simulation makes the validity questionable, especially when measuring against one of the main objectives: the simulations the robot runs should be done on-board the robot.
    \item \emph{Increased scenario complexity and variety} - The current scenario example is a fairly simple example, potentially calculated without machine learning. Scenarios of increasing complexity should be developed to both test the capacity of the RISK system, and being able to test the explainability of the resulting recordings.
    \item \emph{Natural Language Explanations} - Currently only the visuals are created from recording the hypothetical simulations. Based on simulations conducted a language explanation should be generated. This may assist in clearing up misunderstandings, increase accessibility, and mean the robot will not require a display.
\end{itemize}

\section{Acknowledgements}
This work was conducted with support of the Centre for Artificial Intelligence, Robotics and Human-Machine Systems (IROHMS) operation C82092 and partially funded by the European Regional Development Fund (ERDF) through the Welsh Government.

\bibliographystyle{ACM-Reference-Format}
\bibliography{lit}

\end{document}